\documentclass[runningheads]{llncs}
\usepackage{graphicx}
\begin{document}
\title{Multi-Scale Structure-Aware Network for Human Pose Estimation}
%
%
\author{Lipeng Ke\inst{1} \and
Ming-Ching Chang\inst{2} \and
Honggang Qi\inst{1} \and
Siwei Lyu\inst{2}
}
\authorrunning{L. Ke, M. Chang, H. Qi, S. Lyu}
%
\institute{University of Chinese Academy of Sciences, Beijing, China \and
University at Albany, State University of New York, NY, USA 
\email{kelipeng15@mails.ucas.ac.cn\, mchang2@albany.edu \\  hgqi@ucas.ac.cn\, slyu@albany.edu}}
\maketitle              
\begin{abstract}
We develop a robust multi-scale structure-aware neural network for human pose estimation. This method improves the recent deep conv-deconv hourglass models with four key improvements: (1) {\em multi-scale supervision} to strengthen contextual feature learning in matching body keypoints by combining feature heatmaps across scales, (2) {\em multi-scale regression network} at the end to globally optimize the structural matching of the multi-scale features, (3) {\em structure-aware loss} used in the intermediate supervision and at the regression to improve the matching of keypoints and respective neighbors to infer a higher-order matching configurations, and (4) a {\em keypoint masking training} scheme that can effectively fine-tune our network to robustly localize occluded keypoints via adjacent matches. 
Our method can effectively improve state-of-the-art pose estimation methods that suffer from difficulties in scale varieties, occlusions, and complex multi-person scenarios. 
This multi-scale supervision tightly integrates with the regression network to effectively {\em (i)} localize keypoints using the ensemble of multi-scale features, and {\em (ii)} infer global pose configuration by maximizing structural consistencies across multiple keypoints and scales.
The keypoint masking training enhances these advantages to focus learning on hard occlusion samples. 
Our method achieves the leading position in the MPII challenge leaderboard among the state-of-the-art methods.
\end{abstract}
%

\section{Introduction}

Human pose estimation refers to the task of recognizing postures by localizing body keypoints (head, shoulders, elbows, wrists, knees, ankles, {\em etc.}) from images. We focus on the problem of single-person pose estimation from a single RGB image with the input of a rough bounding box of a person, while the pose and the activity of the person can be arbitrary. The task is challenging due to the large variability of human body appearances, lighting conditions, complex background and occlusions, body physique and posture structures of the activities performed by the subject.  The inference is further sophisticated when the case extends to multi-person scenarios. 

\begin{figure*}[t]
\centerline{
\includegraphics[width=0.85\linewidth]{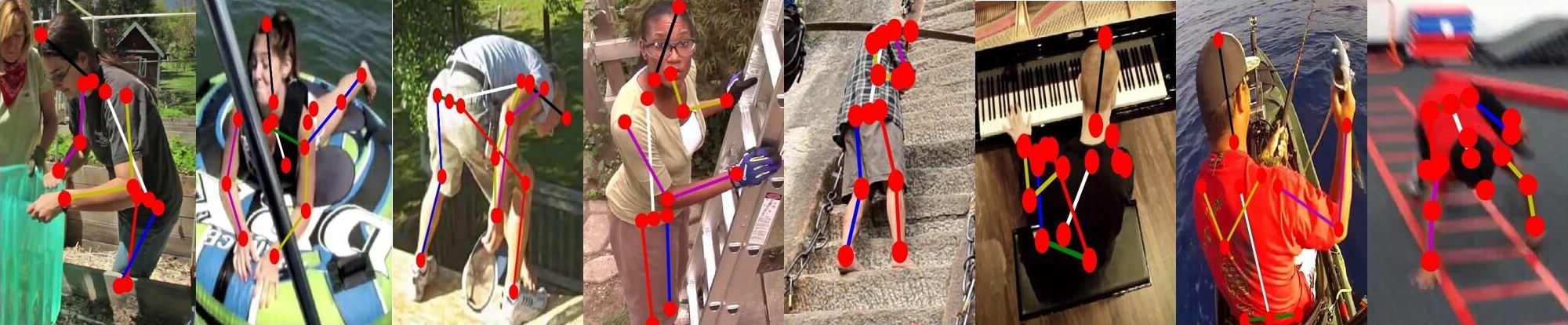}
}
\centerline{
\includegraphics[width=0.85\linewidth]{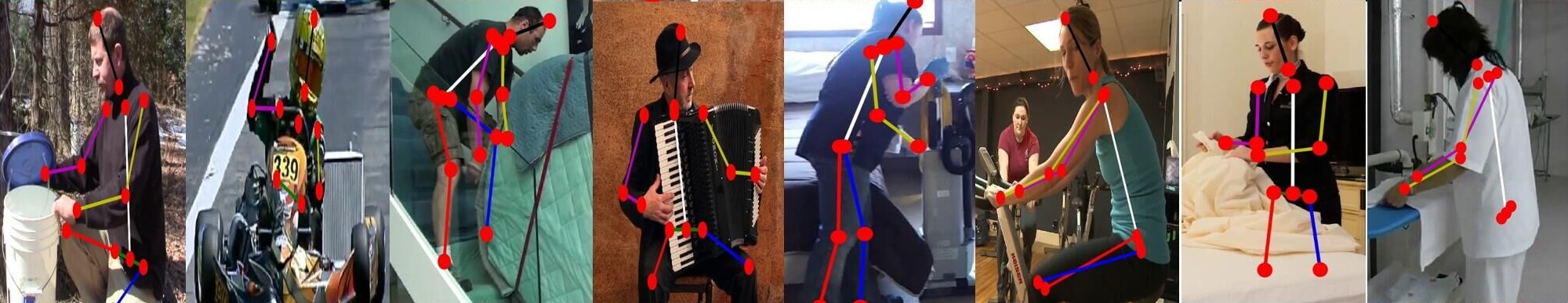}
}
\centerline{
\includegraphics[width=0.85\linewidth]{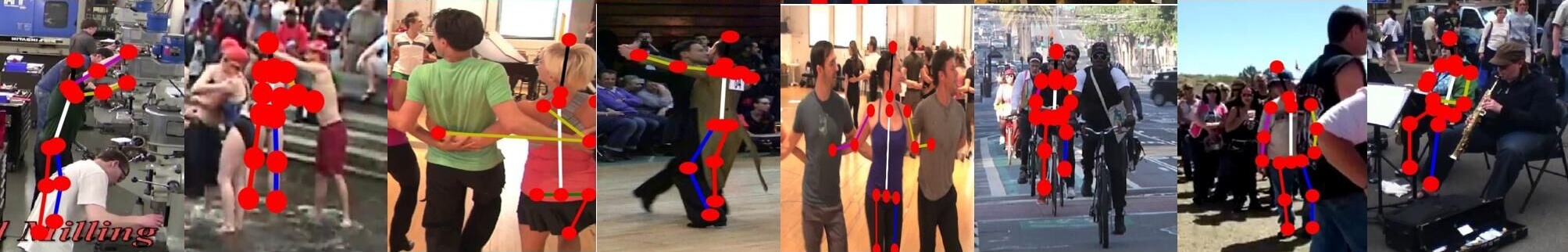}
}
\caption{\em \small 
State-of-the-art pose estimation networks face difficulties in
diverse activities and complex scenes, which can be organized into three challenges:
(top row) large scale varieties of body keypoints in the scenes,
(middle row) occluded body parts or keypoints,
(bottom row) ambiguities in matching multiple adjacent keypoints in crowded scenes.
}
\label{fig:pose_challenge}
\end{figure*}

\begin{figure*}[t]
\centerline{
  \includegraphics[width=\linewidth]{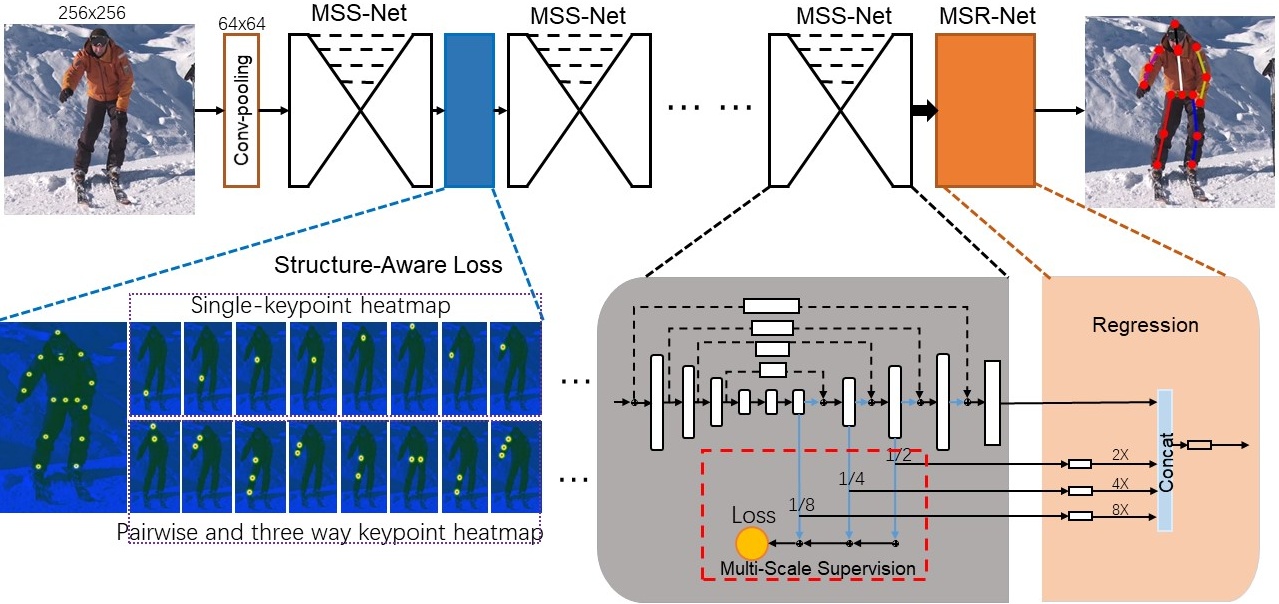}
}
\caption{\em \small 
The proposed network consists of three components: 
{\em (i)} {\em multi-scale supervision network} (MSS-net, $\S$ \ref{sec:multiscale:supervision}),
{\em (ii)} {\em multi-scare regression network} (MSR-net, $\S$ \ref{sec:global:regression:network}),
and 
{\em (iii)} intermediate supervision using the {\em structure-aware loss} ($\S$ \ref{sec:structure:aware:loss}). 
The whole network pipeline is fine-tuned using the {\em keypoint masking training} scheme ($\S$ \ref{sec:keypoint:mask}).
}
\label{fig:pipeline}
\end{figure*}

Human pose estimation has been studied extensively \cite{liu2015a}.
Traditional methods rely on hand-craft features \cite{Bourdev:Malik:Poselet:ICCV2009,Charles:BBC:Pose:BMVC2013,Cherian:MixingBodyPart:CVPR2014,Sapp:Taskar:MODEC:CVPR2013,chang_bmvc15}. With the prosperity of Deep Neural Networks (DNN), Convolutional Neural Networks (CNN) \cite{Toshev:DeepPose:CVPR2014,pfister2015flowing,Tompson:Unified:Pose:NIPS2014,Chu:Structured:Learning:Pose:CVPR2016,wei2016convolutional}, in particular the {\em hourglass} models \cite{newell2016stacked} and their variants \cite{chu2017multi,yang2017learning} have demonstrated remarkable performance in human pose estimation. The repeated bottom-up and top-down processing within the hourglass modules can reliably extract posture features across scales and viewing variabilities, and thus effectively localize body keypoints for pose estimation.

Although great progress has been made, state-of-the-art DNN-based pose estimation methods still suffer from several problems (Fig.~\ref{fig:pose_challenge}):

\noindent (1) {\bf Scale instability:} Slight perturbation of the input bounding box from the person detector (such as the SSD \cite{liu2016ssd}) can cause abrupt changes in the pose estimation, due to the influence of such dominating scales. Such scale instability causes unreliable pose estimations, and even the latest hourglass methods (\cite{chu2017multi,yang2017learning}) tend to overfit body keypoints in a particular scale (out of all scales in the deconv pyramid), which results in a domination of a single scale. Current practice to handle this scale instability ({\em e.g.} widely used in the MPII pose estimation challenge \cite{andriluka14cvpr}), is to repeatedly performing pose estimations in multiple trials of various scales, and output the result with the highest score. This clearly shows the lack of a consistent scale representation in limitations of the existing methods. This will be addressed in this work in $\S$~\ref{sec:multiscale:supervision} and $\S$~\ref{sec:global:regression:network}. 
%

\noindent (2) {\bf Insufficient structural priors:} The second issue is how to effectively incorporate the structure of human body as priors in the deep network for pose estimation. 
Such priors can provide key information to solve challenges of pose estimation in real-world scenarios with complex multi-person activities and cluttered backgrounds, where body keypoint occlusions and matching ambiguities are the bottlenecks.
In these challenge cases, accurate keypoint localization is not the only factor for successful pose estimation, as there will be questions on how best to associate the keypoints (invisible, or multiple visible ones among possibilities) to infer the global pose configuration. Known body structural priors can provide valuable cues to infer the locations of the hidden body parts from the visible ones.
We propose to model the skeleton
with an {\em intermediate structural loss} ($\S$~\ref{sec:structure:aware:loss}) and through the use of a global regression network at the end ($\S$~\ref{sec:global:regression:network}).
We further develop a keypoint masking scheme to improve the training of our network on challenging cases of severely occluded keypoints  ($\S$~\ref{sec:keypoint:mask}).

In this paper, we propose a holistic framework to effectively address the drawbacks in the existing state-of-art hourglass networks. Our method is based on two neural networks: the {\bf multi-scale supervision network} (MSS-net) and the {\bf multi-scale regression network} (MSR-net).

In MSS-net, a layer-wise loss term is added at each deconv layer to allow explicit supervision of scale-specific features in each layer of the network. This multi-scale supervision enables effective learning of multi-scale features that can better capture local contextual features of the body keypoints. In addition, coarse-to-fine deconvolution along the resolution pyramid also follows a paradigm similar to the {\em attention mechanism} to focus on and refine keypoint matches.
The MSR-net takes output from multiple stacks of MSS-nets to perform a global keypoint regression by fusing multiple scales of keypoint heatmaps to determine the pose output.

In addition to the MSS-net and MSR-net which can jointly learn to match keypoints across multiple scales of features, we explicitly match connected keypoint pairs based on the connectivity and structure of human body parts. For example, the connectivity from the elbow to the lower-arm and to the wrist can be leveraged in the inference of an occluded wrist, when the elbow and lower-arm are visible. Hence, we add a {\em structure-aware loss} aims to improve the capacities of the current deep networks in modeling structural priors for pose estimation. This structure loss improves the estimations of occluded keypoints in complex or crowded scenarios.  Lastly, our {\em keypoint masking training} scheme serves as an effective data augmentation approach to enhance the learning of the MSS-net and MSR-net together, to better recognize occluded poses from difficult training samples.

The main contributions of this paper can be summarized as follows:
\begin{itemize} \itemsep 0em
\item We introduce the {\bf multi-scale supervision network} (MSS-net) together with the {\bf multi-scale regression network} (MSR-net) to combine the rich multi-scale features to improve the robustness in keypoint localization by matching features across all scales.
\item Both the MSS-net and MSR-net are designed using a {\bf structure-aware loss} to explicitly learn the human skeletal structures from multi-scale features that can serve a strong prior in recovering occlusions in complex scenes.
\item We propose a {\bf keypoint masking training} scheme that can fine-tune our network pipeline by generating effective training samples to focus the training on difficult cases with keypoint occlusions and cluttered scenes. Fig. \ref{fig:pipeline} summarizes our multi-scale structure-aware network pipeline.
\end{itemize}
Experimental evaluations show that our method achieves state-of-the-art results on the MPII pose challenge benchmark.

\section{Related Work}

Image-based human pose estimation has many applications, for a comprehensive survey, see \cite{liu2015a}. Early approaches such as the histogram of oriented gradients (HOG) and deformable parts model (DPM) rely on hand-craft features and graphical models \cite{Bourdev:Malik:Poselet:ICCV2009,Lafferty01conditionalrandom,Charles:BBC:Pose:BMVC2013,Cherian:MixingBodyPart:CVPR2014,Sapp:Taskar:MODEC:CVPR2013,chang_bmvc15}. These methods suffer from the limited representation capabilities and are not extensible to complex scenarios.

Pose estimation using deep neural networks (DNN) \cite{Toshev:DeepPose:CVPR2014,pfister2015flowing,Tompson:Unified:Pose:NIPS2014,Chu:Structured:Learning:Pose:CVPR2016,wei2016convolutional} has shown superior performance in recent years, due to the availability of larger training datasets and powerful GPUs.  DeepPose developed by Toshev {\em et al.} [4] was an early attempt to directly estimate the postural keypoint positions from the observed image. Tompson {\em et al.} \cite{Tompson:Unified:Pose:NIPS2014} adopted the {\em heatmap} representation of body keypoints to improve their localization during training. 
A Markov random field (MRF) inspired spatial model was used to estimate keypoint relationship. 
Chu {\em et al.} \cite{chu2016structured} proposed a transform kernel method to learn the inter-relationships between highly correlated keypoints using a bi-directional tree. 

Recently, Wei {\em et al.} \cite{wei2016convolutional} used very deep sequential conv-deconv architecture with large receptive fields to directly perform pose matching on the heatmaps. They also enforced {\em intermediate supervision} between conv-deconv pairs to prevent gradient vanish, thus a very deep network became feasible, and the deeper network can learn the keypoints relationship with lager receptive field. 
The hourglass module proposed by Newell {\em et al.} \cite{newell2016stacked} is an extension of Wei {\em et al.} with the addition of residual connections between the conv-deconv sub-modules. The {\em hourglass} module can effectively capture and combine features across scales.
Chu {\em et al.} \cite{chu2017multi} adopted stacked hourglass networks to generate attention maps from features at multiple resolutions with various semantics. Yang {\em et al.} \cite{yang2017learning} designed a pyramid residual module (PRM) to enhance the deep CNN invariance across scales, by learning the convolutional filters on various feature scales.

State-of-the-art DNNs for pose estimation are still limited in the capability of modeling human body structural for effective keypoint matching.
Existing methods rely on a brute-force approach by increasing network depth to implicitly enrich the keypoint relationship modeling capability. A major weakness in this regard is the ambiguities arising from the occlusions, clutter backgrounds, or multiple body parts in the scene. In the MPII pose benchmark \cite{andriluka14cvpr}, many methods \cite{Chu:Structured:Learning:Pose:CVPR2016,wei2016convolutional,newell2016stacked,chu2017multi,yang2017learning} rely on repeating their pose estimation pipeline multiple times in various scales, in order to improve performance by a small margin using averaging of results. This indicates the lack of an effective solution to handle scale and structural priors in the modeling.

\section{Method}
\label{sec:method}

%
%

Our multi-scale structure-aware network consists of two types of subnetworks: the {\em multi-scale supervision network} (MSS-net), which can be repeated for multiple stack, and the {\em multi-scale regression network} (MSR-net) at the end, see Fig. \ref{fig:pipeline}.
Specifically, MSS-net is based on the conv-deconv hourglass module \cite{newell2016stacked} trained with multi-scale loss supervision. 
The MSR-net performs a final pose structural regression by matching multi-scale keypoint heatmaps and their high-order associations.
Both the MSS-net and the MSR-net share a common  {\em structure-aware loss} function, which is designed to ensure effective multi-scale structural feature learning. 
The training of the whole pipeline is fine-tuned using the {\em keypoint masking training} scheme to focus on learning hard samples.

\begin{figure}[t]
\centerline{
\includegraphics[width=0.85\linewidth]{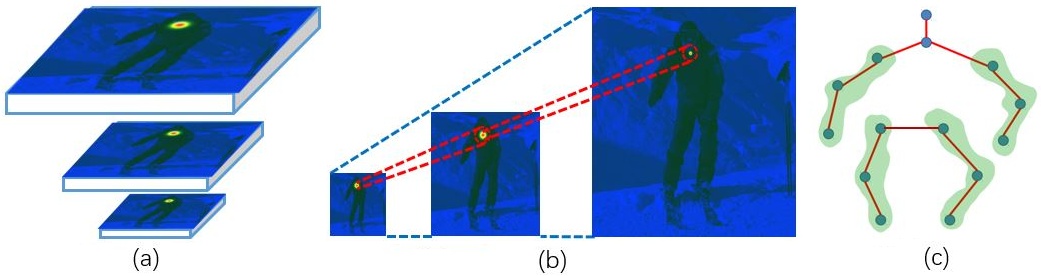}
}
\caption{\em \small 
In the multi-scale supervision network, the refinement of keypoint localization in up-sampling resolution works in analogy to the `attention' mechanism used in the conventional resolution pyramid search.
(a) shows the multi-scale heatmaps of the keypoint of the thorax.
(b) shows the refinement of the keypoint heatmaps during the deconv up-sampling, where the location of the thorax is refined with increased accuracy. 
(c) shows our human skeletal graph with the visualization of keypoint connectivity links.
}
\label{fig:res:pyramid:attention}
\end{figure}


We describe two key observations that motivates the design of our method. 
First, the conv-deconv hourglass stacks capture rich features for keypoint detection across large variability in appearances and scales.
However, such capability is very sensitive to a particular scale in the multi-scale pyramid, and lacks of a robust and consistent response across scales. 
This leads us to add explicit layer-wise supervisions to each of the deconv layer in the training of our MSS-net.

Secondly, the output of the MSS-net hourglass model is a set of heatmaps, and each heatmap corresponds to the location likelihood of each body keypoint (elbows, wrists, ankles, knees, {\em etc}). 
To train the MSS-net, the heatmaps are supervised against the ground-truth body keypoint heatmaps that are typically generated using 2D Gaussian blurring.
At the testing of the MSS-net for pose estimation, the obtained heatmaps are mostly non-Gaussian, which variate according to the gesture of the subject. 
A key deficiency in the original hourglass model \cite{newell2016stacked} is that each keypoint heatmap is estimated {\em independently}, such that the relationship between the keypoints are not considered. In other words, structural consistency among detected keypoints are not optimized. 

To ensure structural consistency in the pose estimation pipeline, we introduce the {\em structure-aware loss} in between the MSS-net hourglass modules that serve as the purpose of intermediate supervision, to better capture the adjacency and associations among the body keypoints. 
The structure-aware loss is also used in the MSR-net at the end of the pipeline, to globally oversee all keypoint heatmaps across all scales.
This way a globally consistent pose configuration can be inferred as the final output. 
The MSR-net regression not only matches individual body keypoints (first-order consistency), but also matches pairwise consistencies among adjacent keypoints (second-order consistency). To illustrate, the co-occurrence of a matching pair between a hand/leg {\em w.r.t.} the head/torso with high confidence should provide stronger hypothesis, in comparison to the separated, uncorrelated individual matches for the final pose inference. 
The MSR-net is trained to perform such optimization across all body keypoints, all scales of features, and all pairwise correlations in a joint regression.

\subsection{Multi-Scale Supervision Network}
\label{sec:multiscale:supervision}

The {\em multi-scale supervision network} (MSS-net) is designed to learn deep features across multiple scales.
We perform multiple layer-wise supervision at each of the deconv layers of the MSS-net, where each layer corresponds to a certain scale.
The gray box at the bottom of Fig.~\ref{fig:pipeline} depicts the MSS-net architecture.

Multi-scale supervision is performed by calculating the residual at each deconv layer using the corresponding down-sampled ground-truth heatmaps in the matching scale ({\em e.g.}, 1/2, 1/4, 1/8 down-sampling). 
Specifically, to make equal the feature map dimensions in order to compute the residual at the corresponding scale, we use an 1-by-1 convolutional kernel for dimension reduction, to convert the high-dimensional deconv feature maps into the desired number of features, where the number of reduced dimension matches the number of body keypoints (which is also the number of heatmaps). 
On the other hand, the ground-truth keypoint feature map is down-sampled to match the corresponding extracted keypoint heatmap at each scale to compute the residual.

The multi-scale supervision network localizes body keypoints in a way similar to the `attention model' \cite{Zhao2017Diversified} used in the conventional resolution pyramid for image search.
The activation areas in the low-res heatmap can provide guidance of the location refinement in the subsequent high-res layers, see Fig. \ref{fig:res:pyramid:attention}.


\begin{figure}[t]
\centerline{
\includegraphics[width=0.85\linewidth]{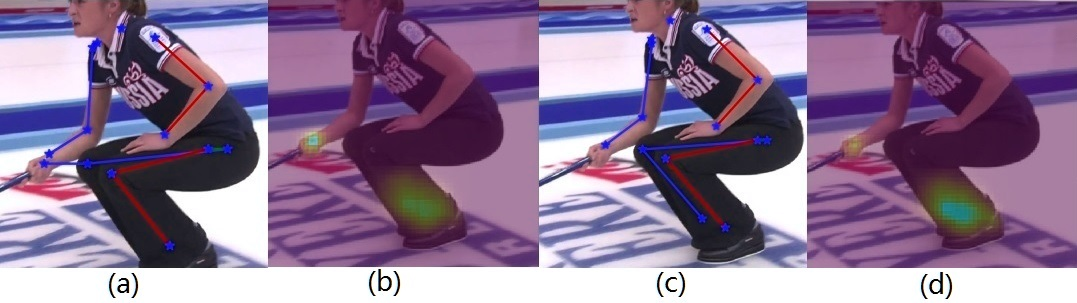}
}
\caption{\em \small 
\textbf{Muti-scale keypoint regression} to disambiguate multiple peaks in the keypoint heatmaps. 
(a-b) shows an example of (a) keypoint prediction and (b) heatmap from the MSS-net hourglass stacks, which will be fed into the MSR-net for regression. 
(c-d) shows (c) the output keypoint locations and (d) heatmap after regression. 
Observe that the heatmap peaks in (d) are more focused compared to (b). 
}
\label{fig:keypoint_regression}
\end{figure}



We describe the loss function $L_{MS}$ to train the multi-scale supervision network.
The loss $L_{MS}$ is defined by summing the $L_2$ loss from the heatmaps of all keypoints across all scales, similar to the multi-scale loss function in \cite{wei2016convolutional,newell2016stacked}.
To detect the $N=16$ keypoints (head, neck, pelvis, thorax, shoulders, elbows, wrists, knees, ankles, and hips), $N$ heatmaps are generated after each conv-deconv stack. 
The loss at the $i$-th scale compares the predicted heatmaps of all keypoints against the ground-truth heatmaps at the matching scale:
%
\begin{equation}
L_{MS}^i = \frac{1}{N}\sum_{n=1}^{N}\sum_{x,y}||P_n(x,y) - G_n(x,y)||_2,
\label{eq:l2:loss}
\end{equation}
where $P_n(x,y)$ and $G_n(x,y)$ represent the predicted and the ground-truth confidence maps at the pixel location $(x,y)$ for the $n$-th keypoint, respectively. 

In standard dataset the ground-truth poses are provided as the keypoint locations. We follow the common practice for ground-truth heatmap generation as in Tompson {\em et al.} \cite{tompson2014joint}, where the $n$-th keypoint ground-truth heatmap $G_n(x,y)$ is generated using a 2D Gaussian centered at the keypoint location $(x,y)$, with standard deviation of 1 pixel.
Fig. \ref{fig:pipeline} (bottom left, first row) shows a few examples of the ground-truth heatmaps for a certain keypoints.

\subsection{Multi-Scale Regression Network}
\label{sec:global:regression:network}

%

We use a {\em fully convolutional} multi-scale regression network (MSR-net) after the MSS-net conv-deconv stacks to globally refine the multi-scale keypoint heatmaps to improve the structural consistency of the estimated poses. 
The intuition is that the relative positions of arms and legs {\em w.r.t.} the head/torso provide useful action priors, which can be learned from the regression network by considering feature maps across all scales for pose refinement. The MSR-net takes the multi-scale heatmaps as input, and match them to the ground-truth keypoints at respective scales. This way the regression network can effectively combine heatmaps across all scales to refine the estimated poses.


The multi-scale regression network jointly optimizes the global body structure configuration via determining connectivity among body keypoints based on the mutli-scale features. This can be viewed as an extension to the work of the Convolutional Part Heatmap Regression \cite{bulat2016human}, which only considers keypoint heatmap regression at the scale of the input image. The input image with the keypoint heatmaps can be seen as an attention method and provide larger resolution. 
In this case, the multi-scale regression network learns a scale-invariant and attention based structure model, thus provide better performance. Moreover, our multi-scale regression network optimizes the structure-aware loss, which matches individual keypoints as well as the higher-order association (pairs and triplets of keypoints) in estimating pose.
The output from the multi-scale regression network is a comprehensive pose estimation that considers pose configurations across multiple feature scales, multiple keypoint associations, and high-order keypoint associations.


Fig.~\ref{fig:keypoint_regression} shows the efficacy of the multi-scale, high-order keypoint regression performed in the MSR-net. The MSR-net works hand-in-hand with the MSS-net to explicitly model the high-order relationship among body parts, such that posture structural consistency can be maintained and refined.


\begin{figure}[t]
\centerline{
\includegraphics[width=0.85\linewidth]{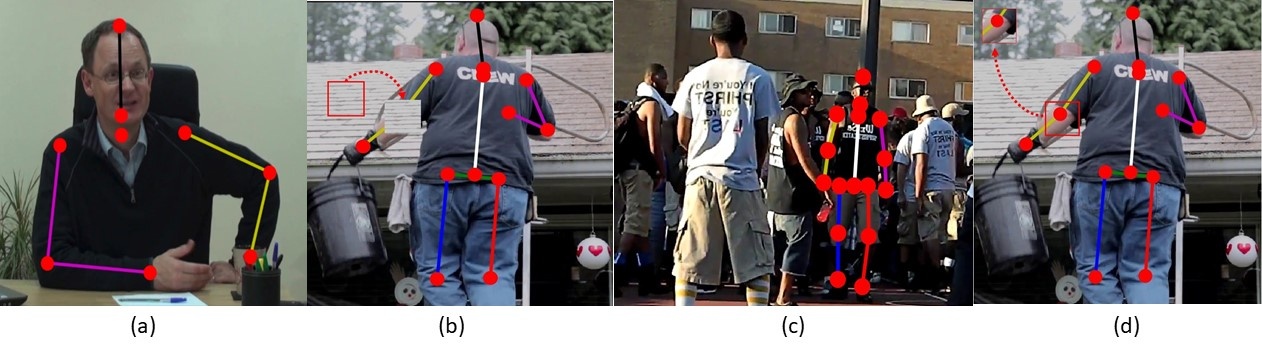}
}
\caption{\em \small {\bf Keypoint masking} to simulate the hard training samples. (a) is a common case in human pose estimation, the keypoint (left-wrist) is occluded by an object, but it can be estimated from the limbs. (c) is another difficult case, where the nearby persons' keypoint can be mismatched to the target person. Thus there are two kind of keypoint masking, (b) is the background keypoint masking which crop a background patch and paste on a keypoint to simulate the keypoint invisible, (d) is the keypoint duplicate masking which crop a keypoint patch and paste on another keypoint to simulate multi-person or multi-peak keypoint heatmap.}
\label{fig:keypoint_masking}
\end{figure}

\subsection{Structure-Aware Loss}
\label{sec:structure:aware:loss}
It has been observed that deeper hourglass stacks lead to better pose estimation results \cite{newell2016stacked}.
As the depth of hourglass stacks increases, {\em gradient vanishing} becomes a critical issue in training the network, where {\em intermediate supervision} \cite{wei2016convolutional,newell2016stacked,chu2017multi,yang2017learning} is a common practice to alleviate gradient vanishing. 

To this end, we design a structure-aware loss function following a graph to model the human skeletal structure. Specifically we introduced a {\em human skeletal graph} ${\cal S}$ (See Fig. \ref{fig:res:pyramid:attention}(c) for a visualization of the human skeletal graph.) to define the structure-aware loss. 
Each node $S_n \in {\cal S}$ represent a body keypoint of the human skeleton and its connected keypoints, $n \in \{1, ..., N\}$.
The {\em structure-aware loss} at the $i$-th scale is formally defined as:
\begin{equation}
L_{SA}^i = \frac{1}{N}\sum_{n=1}^{N}||P_n^i - G_n^i||_2 + \alpha \sum_{i=1}^{N}||P_{S_n}^i - G_{S_n}^i||_2.
\label{structure_loss}
\end{equation}
The first term is the multi-scale supervision loss $L_{MS}^i$ in Eq.\ref{eq:l2:loss} that represents individual keypoint matching loss. 
The second term represents the structural matching loss, where $P_{S_n}$ and $G_{S_n}$ are the combination of the heatmaps from individual keypoint $n$ and its neighbors in graph ${\cal S}$. Hyperparameter $\alpha$ is a weighing parameter balancing the two terms.

Fig. \ref{fig:pipeline} (bottom left) shows a breakdown visualization of how our skeleton-guided structure-aware loss is calculated in traversing the keypoints and their relationships according to ${\cal S}$. The top row in the sub-figure shows the intermediate loss defined on individual keypoints ({\em e.g.}, the right ankle, knee, hip, pelvis, thorax, head, wrist, elbow) as used in \cite{wei2016convolutional,newell2016stacked}.
The bottom row shows our structure-aware loss defined for a set of connected keypoints.

We consider connected keypoints, {\em e.g.}, head-thorax, shoulder-elbow, wrist-elbow, hip-knee, hip-hip, knee-ankle, in the bottom sub-figure of Fig. \ref{fig:pipeline}.
Because that the elbows and knees has additional physical connections (to the shoulders and wrists, and the hips and ankles, respectively), the structure-aware loss in these two joints are three-way to include a triplet of connected keypoints, {\em e.g.}, hip-knee-ankle, shoulder-elbow-wrist as in Fig. \ref{fig:pipeline}. In all cases, the list of structurally connected keypoints is empirically determined according to the human skeletal graph ${\cal S}$, such that the loss can better capture the physical connectivity of the keypoints in the human body to obtain structural priors.

The structure-aware loss is used at two places in our network:
(1) in-between the MSS-net stacks as a means of {\em intermediate supervision} to enforce structural consistency while localizing keypoints; and 
(2) in the MSR-net to find a globally consistent pose configuration.



%


\subsection{Keypoint Masking Training}
\label{sec:keypoint:mask}

In the case of multi-person scenarios, more than one possible body keypoints can co-exist in the view. In occluded case, no keypoint can be visible observed. To tackle these challenging scenarios, we develop a novel {\em keypoint masking} data augmentation scheme to increase training data to fine-tune our networks.  

Specifically, occlusion of key points is an aspect that strongly affects the performance of pose estimation methods. As shown in Fig. \ref{fig:keypoint_masking} (a), the left wrist of the person is occluded by the mug, however the occluded wrist indeed can be estimated by visible connected keypoint(left elbow) o r the libs connecting wrist and elbow. Another difficult case is where there is another person nearby, {\em e.g.} in Fig. \ref{fig:keypoint_masking} (c) that several people standing closely. In this case, the pose estimator may easily take the nearby person's keypoint as its own keypoint. 
One drawback of training the network using the original training set is that there usually exists insufficient amount of examples that contains the occlusion cases to train a deep network for accurate keypoint detection/localization. 
Conventional data augmentation method, such as the popular horizontally flipping, random crops and color jittering in classification, are not helpful in this case.
 
We propose a keypoint masking method to address this problem by copying and pasting body keypoint patches on the image for data augmentation. 
The main idea is to generate keypoint occluded training samples as well as the artificially inserted keypoints, such that the network can effectively improve its learning on these extreme cases. This data augmentation is easily doable from the known ground-truth keypoint annotations.

Specifically, we introduce two types of keypoint/occlusion sample generation methods: 
(1) As shown in Fig. \ref{fig:keypoint_masking} (b), we copy a background patch and put it onto a keypoint to cover it, in order to simulate a keypoint occlusion.
This kind of sample is useful for the learning of occlusion recovery. 
(2) As shown in Fig. \ref{fig:keypoint_masking} (d), we copy a body keypoint patches and put it onto a nearby background, in order to simulate the multiple existing keypoints, the case that mostly occurs in multi-person scenarios. 
Since this data augmentation results in multiple identical keypoint patches, the solution to a successful pose estimation must rely on some sort of structural inference or knowledge.
It is thus especially beneficial to fine-tune to our global keypoint regression network. 

Overall this keypoint masking strategy can effectively improve the focus of learning on challenge cases, where important body keypoints are purposely masked out or artificially placed at wrong locations. The effect of keypoint masking training in improving both (1) the detection and localization of occluded keypoints and (2) global structure recognition will be evaluated in \S\ref{sec:component:analysis}.

\section{Experiments and Analysis}
\label{sec:exp}


We train and test our model on a workstation with 4 NVIDIA GTX 1080Ti GPUs and two public datasets -- the MPII Human Pose Dataset and Challenge \cite{andriluka14cvpr} and FLIC dataset \cite{Sapp:Taskar:MODEC:CVPR2013}.
The MPII dataset consists of images taken from a wide range of real-world activities with full-body pose annotations. It is considered as the ``de facto'' benchmark for state-of-the-art pose estimation evaluation.
The MPII dataset includes around 25K images containing over 40K subjects with annotated body joints, where 28K subjects are used for training, and the remaining 12k are used for testing. The FLIC dataset consists of $5,003$ selected images obtained from Hollywood movies. The images are annotated on the upper body, where the subjects are mostly facing the camera, thus there exists less keypoint occlusions.

Since the testing annotations for MPII are not available to the public, in our experiments, we perform training on a subset of the original training set, and perform hyper-parameter selection on a separated {\em validation set}, which contains around 3K subjects (that are in the original training set). We also report evaluation results that are reported from the MPII benchmark \ref{sec:eval:accuracy}.
 
\subsection{Implementation}

\subsubsection{Training} 
is conducted on the respective datasets (MPII, FLIC) using the SGD optimizer for 300 epochs. In this work, we use 8 stacks of hourglass modules for both training and testing. The training processes can be decided into three stages: {\em (1)} MSS-Net training, {\em (2)} MSR-Net training, and {\em (3)} the joint training of the MSS-Net and MSR-Net with keypoint masking. 
We use the same data augmentation technique as in the original hourglass work \cite{newell2016stacked} that includes rotation ($+/-$ 30 degrees), and scaling (.75 to 1.25) throughout the training process. 
Due to GPU memory limitation, the input images were cropped and rescaled to 256x256 pixels. For the first stage we train the MSS-Net for 150 epochs, with the initial learning rate to be 5e-4.
The learn rate is reduced by a factor of 5 when the performance dose not improve after 8 epochs. 
We then train the MSR-Net for 75 epochs with the MSS-Net parameters fixed. 
Finally the whole network pipeline is trained for 75 epoch with keypoint masking fine-tuning.  

\subsubsection{Testing}
is performed on both the MPII and FLIC datasets. Since this work focuses on single-person pose estimation, and there often exists multiple subjects in the scene.
We use a conditional testing method --- 
We first test pose estimation in the original scale assuming the subject appears at the image center.
We then check if the detected body keypoint confidence is lower then a specific threshold.
If so, no successful human pose is found.
We then perturb the putative person location, and repeat the pose finding, to see if a refined pose can be found. 
The keypoint confidence thresholds ${\bf \tau}_c$ can be keypoint-dependent, and are empirically determined using the validation set. 
For the case multiple pose estimation test trials are performed, only the results with scores higher than a threshold $\tau_s$ are selected for the fusion of the pose output. The value of $\tau_s$ is also empirically determined from the validation set.
We note that this testing refinement may reduce the testing performance of pose estimation, because the variation of the input (person bounding box) is also considered in the process.


%

\subsection{Evaluation Results}
\label{sec:eval:accuracy}

Evaluation is conducted using the standard Percentage of Correct Keypoints (PCK) metric \cite{tompson2015efficient} which reports the percentage of keypoint detection falling within a normalized distance of the ground truth. For the FLIC evaluation, PCK is set to the percentage of disparities between the detected pose keypoints {\em w.r.t.} the ground-truth after a normalization against a fraction of the torso size. For the MPII evaluation, such disparities are normalized by a fraction of the head size, which is denoted as PCK$^h$.

{\bf FLIC}: Table \ref{table:flic-pck5} summarizes the FLIC results, where our PCK reaches 99.2\% for the elbow, and 97.3\% for the wrist. 
Note that the elbows and wrists are the most difficult parts to localize in the FLIC dataset. Comparison with Newell {\em et al.} \cite{newell2016stacked} demonstrates the improvement of our structure-aware design in the MSS-net and MSR-net in our method.

\begin{figure}[t!]
\centering
\includegraphics[width=0.8\linewidth]{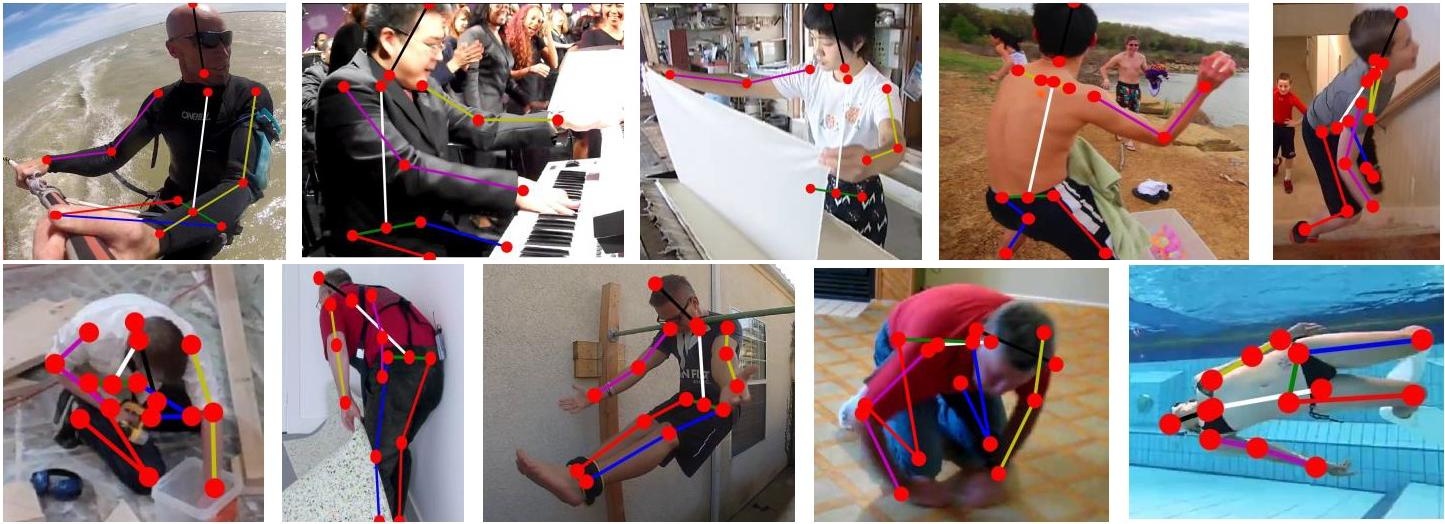}
\includegraphics[width=0.8\linewidth]{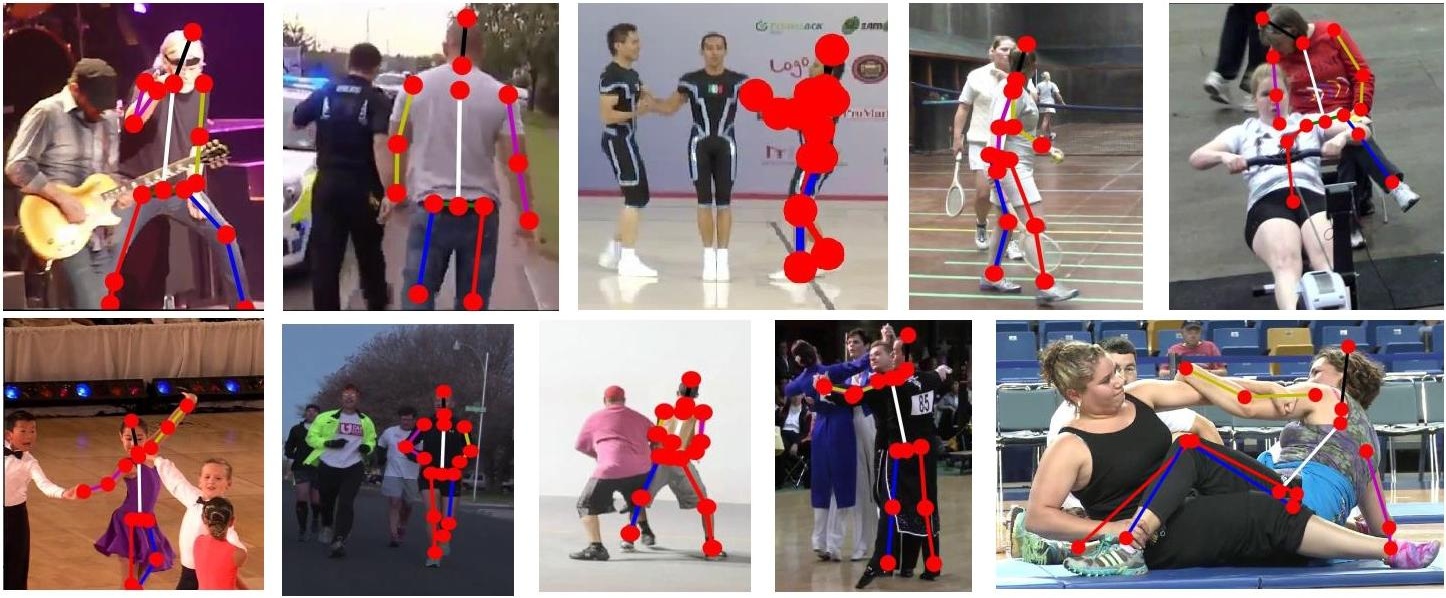}
\includegraphics[width=0.8\linewidth]{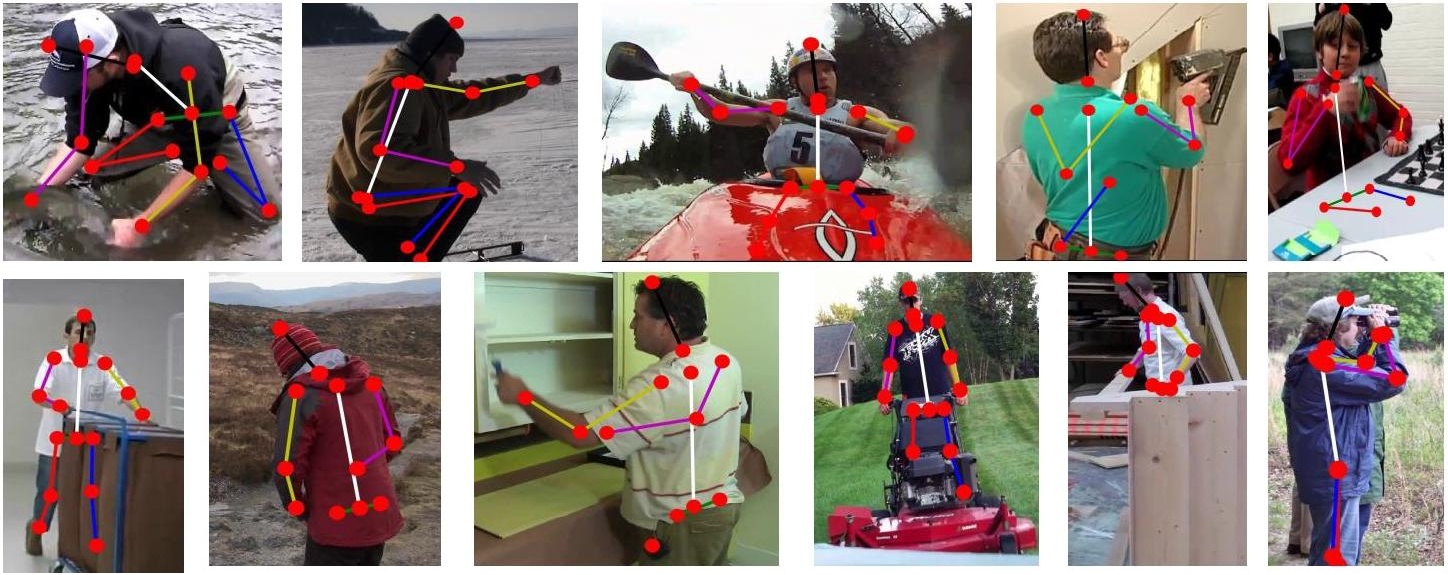}
\caption{\em \small Example of pose estimation results on the MPII dataset using our method. 
(row 1) Examples with significant scale variations for keypoints.
(row 2,3) Examples with multiple persons.
(row 4,5) Examples with severe keypoint occlusions.
}
\label{fig:mpii_example}
\end{figure}

\begin{figure}[t]
\centerline{
  \includegraphics[width=0.85\linewidth]{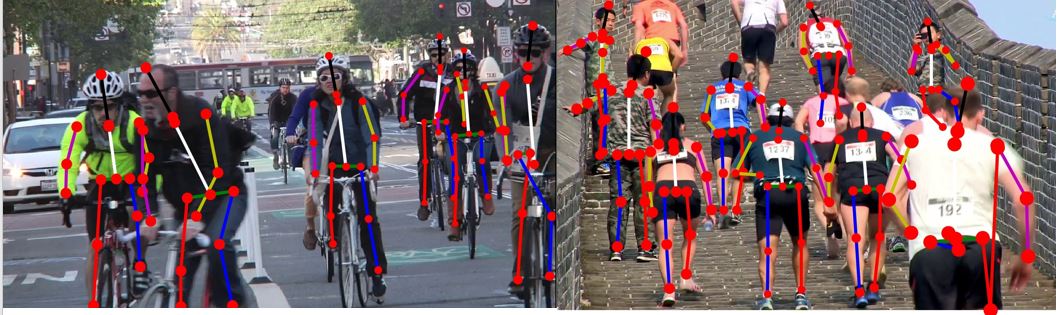}
}
\caption{\small \em Pose estimation results with our method on two very challenging images from the MPII dataset with crowded scene and severe occlusions. Our method can reliably recover complex poses for each targeted person.}
\label{fig:multi-person}
\end{figure}

{\bf MPII}:
Table \ref{table:mpii-pckh5} summarizes the MPII evaluation results. Observe that our method achieves the {\em highest} total score (92.1) and state-of-the-art results across all keypoints on the MPII benchmark as well as the AUC score. In Fig. \ref{fig:mpii_example} we show several pose estimation results on the MPII dataset. In Fig. \ref{fig:multi-person} we show some highly challenging examples with crowded scenes and severe occlusions. In this case, we run our pose estimation on the bounding box of each person, which is provided in the MPII dataset. Our method can extract complex poses for each targeted person, without confusing with other person's poses and in the presence of occlusions.

\begin{table}[t]
	\caption{\em \small 
		Results on the FLIC dataset (PCK = 0.2)
	}
	\begin{center}
		\begin{tabular}{l||*{7}{c}|r}
			\hline
			& Elbow & Wrist\\ \hline
			Tompson {\em et al.} CVPR'15 \cite{Tompson:Unified:Pose:NIPS2014} &93.1 &92.4 \\
			
			Wei {\em et al.} CVPR'16 \cite{wei2016convolutional}  &97.8 &95.0 \\ 
            Newell{\em et al.} ECCV'16 \cite{newell2016stacked} &99.0&97.0\\
            \hline
			Our model& \textbf{99.2} &\textbf{97.3} \\ 
			\hline 
		\end{tabular}
	\end{center}
	\label{table:flic-pck5}
\end{table}

\begin{table*}[t]
\caption{\em \small 
Evaluation results on the MPII pose dataset (PCK$^h$ = 0.5). {Results were retrieved on 03/15/2018.}
}
\begin{center}
\begin{tabular}{l||*{8}{c}|r}
 \hline
 &Head & Shoulder & Elbow & Wrist & Hip & Knee  & Ankle & Total & AUC \\
 \hline
Our method& \bf{98.5}  & \bf{96.8}  & \bf{92.7}  & 88.4  & 90.6  & 89.3 & \bf{86.3} & \bf{92.1} &  63.8 \\
\hline
Chen {\em et al.} ICCV'17 \cite{chen2017adversarial} & 98.1  & 96.5  & 92.5  & \bf{88.5}  & 90.2  & \bf{89.6} & 86.0 & 91.9 & 61.6 \\
Chou {\em et al.} arXiv'17 \cite{DBLP:journals/corr/ChouCC17} & 98.2  & \bf{96.8}  & 92.2  & 88.0  & \bf{91.3}  & 89.1 & 84.9 & 91.8 & 63.9 \\
Chu CVPR'17 \cite{chu2017multi} & \bf{98.5}  & 96.3  & 91.9  & 88.1  & 90.6  & 88.0 & 85.0 & 91.5 & 63.8 \\
Luvizon {\em et al.} arXiv'17 \cite{DBLP:journals/corr/abs-1710-02322} & 98.1  & 96.6  & 92.0  & 87.5  & 90.6  & 88.0 & 82.7 & 91.2 & 63.9 \\
Ning {\em et al.} TMM'17 \cite{DBLP:journals/corr/abs-1710-10192} & 98.1  & 96.3  & 92.2  & 87.8  & 90.6  & 87.6 & 82.7 & 91.2 & 63.6 \\
Newell ECCV'16 \cite{newell2016stacked} & 98.2  & 96.3  & 91.2  & 87.1  & 90.1  & 87.4 & 83.6 & 90.9 & 62.9 \\
Bulat ECCV'16 \cite{bulat2016human} & 97.9  & 95.1  & 89.9  & 85.3  & 89.4  & 85.7 & 81.7 & 89.7 & 59.6 \\
Wei CVPR'16 \cite{wei2016convolutional} & 97.8  & 95.0  & 88.7  & 84.0  & 88.4  & 82.8 & 79.4 & 88.5 & 61.4 \\
Insafutdinov ECCV'16 \cite{insafutdinov2016deepercut} & 96.8  & 95.2  & 89.3  & 84.4  & 88.4  & 83.4 & 78.0 & 88.5 & 60.8 \\
Belagiannis FG'17 \cite{belagiannis2017recurrent} & 97.7  & 95.0  & 88.2  & 83.0  & 87.9  & 82.6 & 78.4 & 88.1 & 58.8 \\

\hline
\end{tabular}
\end{center}
\label{table:mpii-pckh5}
\end{table*}

\subsection{Component Analysis}
\label{sec:component:analysis}

We performed a series of ablation experiments to investigate the effect of individual components in our method. The ablation study is conducted on the validation set \cite{tompson2014joint} of the MPII dataset. 
Note that our method can be reduced to the original hourglass model of Newell {\em et al.} \cite{newell2016stacked} after all newly proposed features are taken out.
Thus, we analysis each proposed network design, {\em i.e.}, the MSS-net, MSR-net, structure-aware loss, and keypoint masking, by comparing against Newell {\em et al.} with a baseline score of 87.1\% at PCK$^h = 0.5$.


{\bf Multi-scale supervision} (MSS-net without structure-aware loss): We first evaluate the effect of the multi-scale supervision along. By adding the multi-scale supervision at the deconv layers of hourglass model \cite{newell2016stacked}, the PCK$^h$ score improve from 87.1\% to 87.6\% and also with a significant computation reduction. This is because the original hourglass method \cite{newell2016stacked} is tested with input images of multiple scales (6 scales in our experiment), while the evaluation of our multi-scale supervision network only need to be tested once in the original scale input. Our method does not require repeated runs and fusion of different scales as post-processing.

{\bf Multi-scale regression} (MSS-net and MSR-net without structure-aware loss): To justify the contribution of multi-scale regression, we evaluate the effect of the second stage of our training pipeline ({\em i.e.} the 
MSR-net after the MSS-net is trained, without keypoint masking fine-tuning). The PCK$^h$ score here is 88.1\% score, which is 0.4\% improvement brought by the multi-scale regression.

{\bf Structure-aware loss} (MSS-net and MSR-net with structure-aware loss): The next in our ablation pipeline is to use structure-aware loss in the training of MSS-net and MSR-net, in comparison to the original loss defined in Eq.\ref{eq:l2:loss}. The PCK$^h$ score we obtained here is 88.3\%, which is a 0.3\% improvement brought by the use of structure-aware loss for training.
 
{\bf Keypoint masking}: After 75 epochs keypoint masking fine-tuning in the MSS-net and MSR-net pipeline with structure-aware loss, we achieve a 88.4\% PCK$^h$ score. The keypoint masking contributes 0.1\% PCK$^h$ improvement in this ablation study.





\section{Conclusion}

We describe an improved multi-scale structure-aware network for human pose estimation.
The proposed multi-scale approach (multi-scale supervision and multi-scale regression) works hand-in-hand with the structure-aware loss design, to infer high-order structural matching of detected body keypoints, that can improve pose estimation in challenging cases of complex activities, heavy occlusions, multiple subjects and cluttered backgrounds.
The proposed keypoint masking training can focus the learning of the network on difficult samples.
Our method achieve the leading position in the MPII challenge leaderboard among the state-of-the-art methods. Ablation study shows the contribution and advantage of each proposed components.




\bibliographystyle{splncs04}
\bibliography{egbib}
%




\end{document}